\documentclass{article}

\usepackage[square,numbers]{natbib}

\usepackage[preprint]{neurips_2020}

\usepackage[utf8]{inputenc} 
\usepackage[T1]{fontenc}    
\usepackage{hyperref}       
\usepackage{url}            
\usepackage{booktabs}       
\usepackage{amsfonts}       
\usepackage{nicefrac}       
\usepackage{microtype}      
\usepackage{graphicx}       
\usepackage{subfigure}
\usepackage{amsmath}        
\usepackage{setspace}
\usepackage[noend]{algpseudocode}
\usepackage{algorithmicx,algorithm}
\usepackage{multirow}
\usepackage{makecell}
\usepackage{array}
\usepackage{float}
\usepackage{bm}

\title{A New Modal Autoencoder for Functionally Independent Feature Extraction}

\author{
    Yuzhu Guo
    \And 
    Kang Pan
    \And
    Simeng Li
    \And 
    Zongchang Han \\
    \And 
    Kexin Wang  \\
    \and
    \text{Department of Automation Sciences and Electrical Engineering} \\
    Beihang University \\
    Beijing, China \\
    \texttt{\{yuzhuguo, lkwxl, lsimeng, zchan, wkx933\}@buaa.edu.cn}\\
    \And 
    Li Li \\
    \text{Department of Automation Sciences and Electrical Engineering} \\
    Beihang University \\
    Beijing, China \\
    \texttt{lililibuaa@126.com }\\
}
\begin{document}

\maketitle

\begin{abstract}
Autoencoders have been widely used for dimensional reduction and feature extraction. Various types of autoencoders have been proposed by introducing regularization terms. Most of these regularizations improve representation learning by constraining the weights in the encoder part, which maps input into hidden nodes and affects the generation of features. In this study, we show that a constraint to the decoder can also significantly improve its performance because the decoder determines how the latent variables contribute to the reconstruction of input. Inspired by the structural modal analysis method in mechanical engineering, a new modal autoencoder (MAE) is proposed by othogonalising the columns of the readout weight matrix. The new regularization helps to disentangle explanatory factors of variation and forces the MAE to extract fundamental modes in data. The learned representations are functionally independent in the reconstruction of input and perform better in consecutive classification tasks. The results were validated on the MNIST variations and USPS classification benchmark suite. Comparative experiments clearly show that the new algorithm has a surprising advantage. The new MAE introduces a very simple training principle for autoencoders and could be promising for the pre-training of deep neural networks.
\end{abstract}

\section{Introduction}
\label{sec1}
The success of machine learning heavily depends on data representation. A good information representation may significantly simplify the consequent classification or regression tasks. Good representation can also improve the transferability by sharing statistical strength across different tasks and model interpretability by extracting abstract and meaningful features \cite{ref50}. How to use representation learning to defend against adversarial attacks is another hot topic \cite{ref66}.  In many machine learning tasks, the raw data are high-dimensional and sensitive to noise, such as images in object recognition. The high dimensional data may cause a series of problems, including high-computational complexity, being prone to over-fitting, low transferability or low interpretability. Feature extraction may contract the main information into a very low dimensional abstract feature space with properly designed mappings.

Feature engineering is a way to learn representations based on human ingenuity and prior knowledge. While much of this feature-engineering work is extremely clever, the hand-engineering approaches can be labor-intensive and may not scale well to new problems \cite{ref50}. Additionally, the extreme goal of artificial intelligence is to solve the problem automatically and to make the application of machine learning as easy as possible. Hence, it would be highly desirable to make learning algorithms less dependent on background knowledge and feature engineering, so that novel applications could be constructed faster and painlessly.

Earlier feature extraction methods include supervised feature extraction such as LDA \cite{ref59} and unsupervised feature extraction such as PCA \cite{ref60}. The major disadvantage of these methods is that the mappings are linear. Kernel methods mapping features into nonlinear feature spaces using flexible and nonparametric basis kernel functions to describe the local behaviour of the data. However, the obtained representations can be data-dependent and the number of kernels can easily explode and lead to over-fitting. Self-supervised feature extraction provides a promising option for feature extraction by minimizing the reconstruction error of the original signals through an autoencoder and readouting the reprehensive features from the bottle-neck. The reduced high-level features usually contain the main information of the original data, with a much sparser representation. 

Autoencoders prove a powerful method for feature extraction and are also often used for the pre-training of deep neural networks\cite{ref61}. Many variations have been proposed to improve the performance of the basic autoencoder (BAE) by introducing regularization terms. Denoising Autoencoders (DAE) improve the robustness of representation by reconstructing original clean input from a deliberately corrupted signal \cite{ref51}. As a result, DAEs can produce more robust features. Additionally, setting up a single-thread DAE is very easy. Sharing a similar motivation, Contractive Autoencoders (CAE) learn robust representation by constraining the Frobenius norm of the Jacobian matrix of the encoder activations with respect to the input \cite{ref62}. CAEs improve the continuity of the encoder, that is, similar inputs have similar encodings. Both DAEs and CAEs force a model to learn how to contract a neighbourhood of inputs into a smaller neighbourhood of outputs. Sparse Autoencoders (SAE) introduce a sparsity penalty on the hidden layer nodes in addition to the reconstruction error \cite{ref63}. Consequently, each hidden unit learns a typical pattern in the input space and the hidden layer nodes keep in a low activation level. With this design, SAE can still discover interesting structures in the data, even if the number of hidden units is large. 

Most of these regularization methods improve the performance by constraining the weights in the encoder part because they directly connect input and hidden nodes. However, a constraint to the decoder can also affect the performance of an autoencoder because the decoder part determines how these latent variables contribute to the reconstruction of the input. Few results that regularise the decoder part of an autoencoder. A new regularization method, named a modal autoencoder (MAE), which constrains the weights of the decoder to produce functionally independent feature, is introduced in this study.

The remainder of this paper is organised as follows: preliminaries of the autoencoder are briefly reviewed after the introduction. Inspired by modal analysis, the new modal autoencoder algorithm is discussed in detail in Section \ref{sec3}. Section \ref{sec4} illustrates the effectiveness of the new MAE method with extensive numerical experiments. Conclusions are finally drawn in Section \ref{sec5}.

\section{Preliminaries}
\label{sec2}

We begin with recalling basic autoencoder (BAE) models and discuss the desired properties of a good autoencoder before presenting our proposed modal autoencoder (MAE) model.

\begin{figure}
  \centering
  \includegraphics[width = 0.5\linewidth]{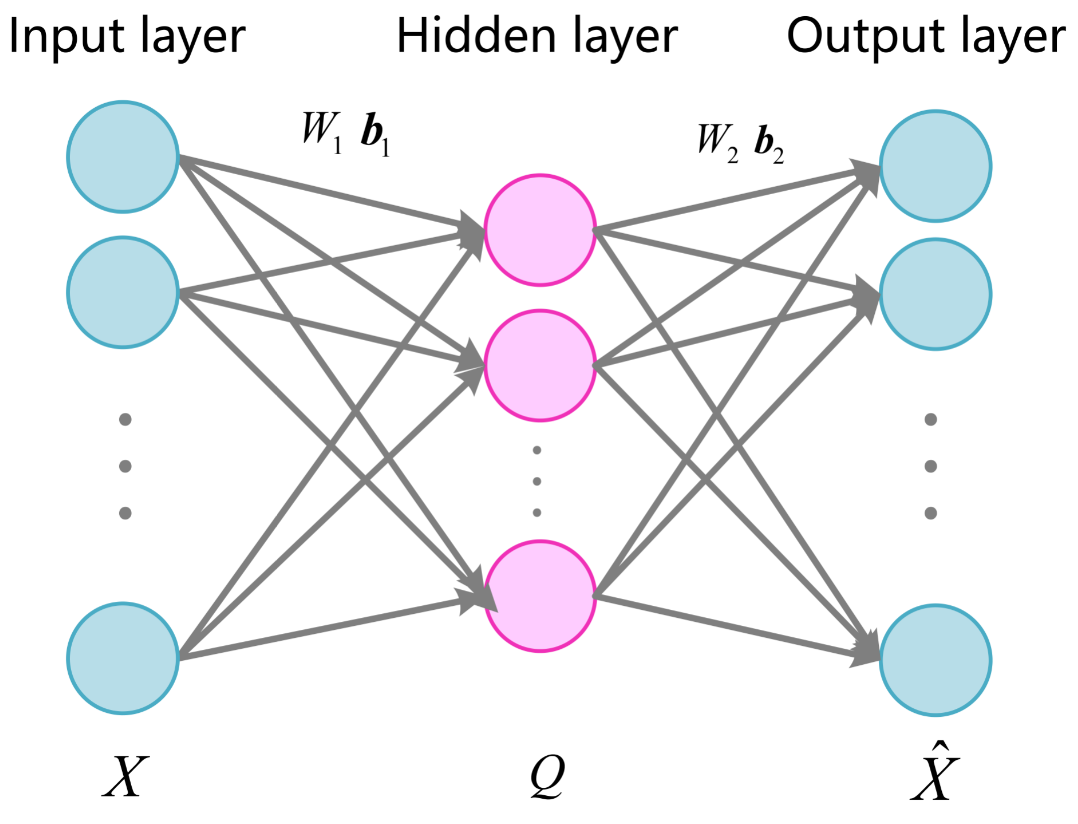}
  \caption{Structure of basic autoencoder model(BAE).}  
  \label{fig1}
\end{figure}

Autoencoders are inspired by the manifold hypothesis and learn lower-dimensional representations directly from high dimensional data. An autoencoder \cite{ref64} is an unsupervised/self-supervised learning artificial neural network, which is capable of learning an efficient representation of the input data. An autoencoder consists of two parts: an encoder and a decoder. Given a $d$ dimensional input $X$ with $n$ samples, the output $Q$ of the encoder represents the input data  $X$  in a transformed space, whose dimension is usually smaller than the dimensions of $X$; the decoder reconstructs $X$ from $Q$, by minimizing the reconstruction error, that is, the difference between the output $\hat{X}$ of decoder and the input $X$ as illustrated in Figure \ref{fig1}.

Specifically, given the input space $ X\in\chi$ and the feature space $Q\in F$, the encoder is a mapping function $f:\chi\to F$, which can be defined as
\begin{equation}
  \begin{split}
    Q=&f\left( X \right)={{s}_{f}}\left( {{W}_{1}}X+{{b}_{1}} \right)
  \label{eq1}
  \end{split}
\end{equation}
where $s_f(\cdot)$ is  a linear or nonlinear activation function; $W_1$ is the weight matrix and $b_1$ is the bias vector.

The decoder is a mapping function $g:F\to\chi$ which is defined as
\begin{equation}
  \begin{split}
    \hat{X}=g\left( Q \right)={{s}_{g}}\left( {{W}_{2}}Q+{{b}_{2}} \right)
  \label{eq2}
  \end{split}
\end{equation}
where $s_g(\cdot)$ denotes the activation function. A linear identity activation function means that the input can be represented by a linear combination of the hidden layer neurons $Q$. $W_2$ is the weight matrix; $ b_2$ is the bias vector, and $\hat{X}$ is reconstruction of the input data $X$.

Training the autoencoder is done to find the optimal mapping pair  $(f,g)$ to minimise the reconstruction error which is measured by the loss function $L$. 

The reconstruction loss function is often defined as the mean squared error loss, that is,
\begin{equation}
  \begin{split}
    L\left( X,\hat{X} \right)=\left\| X-\hat{X} \right\|_{{}}^{2}
  \label{eq3}
  \end{split}
\end{equation}

Thus, the training of the BAE model is to search for solutions to the optimization problem
\begin{equation}
  \begin{split}
    \left(f,g\right)=\underset{\left(f,g\right)}{\mathop{\text{argmin}}}\,L\left(X,g\left[ f\left(X\right)\right]\right)
  \label{eq4}
  \end{split}
\end{equation}
Since the mappings $(f,g)$ can uniquely be determined by the parameters  $\theta=(W_1,b_1,W_2,b_2)$  given the activation functions. The optimization problem can then be rewritten as follows
\begin{center}
  $\theta =\underset{\theta }{\mathop{\text{argmin}}}\,L\left( X,g\left[ f\left( X \right) \right] \right)$
\end{center}
A basic autoencoder without any constraints can generate many representations of the input even when the bottle-neck is small. Hence, many variants of autoencoders have been proposed by integrating prior one to generate features with desired properties, such as the DAE, SAE and CAE . 

Generally, a good set of features are expected to the have following properties: the features should be abstract and invariant; the features should be reusable; and features should be able to disentangle factors of variation of input. A good autoencoder should be able to learn abstract features that are not sensitive to the change of input. Rich types of input configurations can be generated by reusing these features with different weights. Each of the features explains a source of variation of the input and different explanatory factors tend to change independently.  Examples of good representation can often be seen in laws of physics. One example is the modal decomposition of structural vibrations in mechanical or civil engineering. Inspired by the mechanical modal analysis method, a new modal autoencoder is proposed to extract a set of invariant features from observed patterns.  

\section{Modal AutoEncoder}
\label{sec3}
In this section, the modal analysis method in mechanical engineering is briefly reviewed to show how modal analysis is an autoencoder structure and can extract fundamental modes from structural vibrations. A new modal autoencoder is then proposed by transferring some knowledge from mechanical mode decomposition into feature learning.

\subsection{Inspiration from modal analysis}
\label{subsec3.1}
According to structural vibration theory, any vibration of multiple degree of freedom structure can be approximated by a linear superposition of a group of basic modes of vibration \cite{ref56}. These modes represent basic structure information and reveal the inherited information underlying complex vibration signals. For example, the vibration of a board can be decomposed as a linear combination of oscillations with the first four modes.
\begin{figure}
  \centering
  \includegraphics[width = 0.55\linewidth]{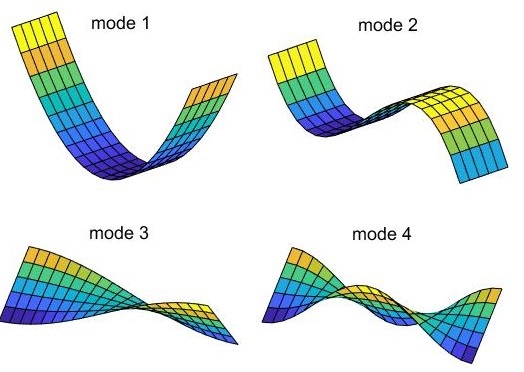}
  \caption{ First four modal shapes of a rectangular board.}  
  \label{fig6}
\end{figure}

Modal analysis provides an invariant description of the structural dynamics. The basic modes and modal shapes are determined by the system structure and do not change with specific vibration signals, which can be observed in an experiment. 

A structural vibration of a $n$ degree of freedom system can be represented in the modal coordinate system as
\begin{equation}
  \begin{split}
    x\left(t\right)=\sum\limits_{i=1}^{m}{{{\psi }_{i}}{{q}_{i}}\left( t \right)}=\Psi q
  \label{eq5}
  \end{split}
\end{equation}
where $x(t)={{[{x}_{1}}(t),{{x}_{2}}(t),\cdots,{{x}_{n}}(t)]^{T}}$ represents the vibration of a structure at $n$  different nodes; $q(t)={{[{q}_{1}}(t),{{q}_{2}}(t),\cdots,{{q}_{m}}(t)]^{T}}$ represents $m$ main vibration modes (oscillations).$\Psi=[ \begin{matrix}{{\psi }_{1}}, {{\psi }_{2}}, \cdots,{{\psi }_{m}}\\\end{matrix}]$ is the modal shape matrix, which determines how each of the modes affects the vibration at different nodes of the structure. 

Modal analysis is used to find the vibration modes and the associated modal shapes from the observation of structural vibrations and uses the invariant modes to study more general structural dynamics. It can be observed that the modal analysis is essentially an autoencoder where the modes are the representational features of vibration and corresponds to hidden nodes in an autoencoder. The study of the mapping from vibration to mode can be considered as the encoding process and the reconstruction of vibration from modes is the decoding process where the modal shape matrix $\Psi$  corresponds to the readout matrix $W_2$. 

One of the most important characteristics of modal decomposition is that the structural modal shapes are orthogonal to each other (with respect to the mass and stiffness matrix), namely, the contribution of each mode is disentangled and each mode affects the vibration independently. This has given us the most inspiration to design extra regularization to disentangle the factors of variation of input in an autoencoder. 

\subsection{Modal autoencoder}
\label{subsec3.2}
Inspired by the mechanical modal analysis, a new autoencoder training principle can be proposed by introducing a constraint to $W_2$, that is, the columns of $W_2$  need to be mutually orthogonal to each other. We call this a functionally independent constraint because the column vector of $W_2$ reflects the functions of each hidden layer node in the reconstruction of the original input. 

A new MAE has the same structure as in Figure \ref{fig1} but is trained with a different loss function
\begin{equation}
  \begin{split}
    {{L}_{MAE}}=\left\| X-\hat{X} \right\|_{{}}^{2}+{{\lambda }_{m}}\left\| W_{2}^{T}M{{W}_{2}}-I \right\|_{F}^{2}
  \label{eq6}
  \end{split}
\end{equation}
which is composed of two parts: the reconstruction error and an additional regularization term. Thus, training the MAE is to solve the minimization problem
\begin{equation}
  \begin{split}
    \theta =\underset{\theta }{\mathop{\text{argmin}}}\,{{L}_{MAE}}
  \label{eq7}
  \end{split}
\end{equation}
where $\theta=(W_1,b_1,W_2,b_2)$ is the parameters to be optimised; $M$ is a constant square matrix; $I$ is the identity matrix;  $\|\cdot\|_F$ represents the Frobenius norm;$\lambda_m$ is a regularization parameter which balances the reconstruction loss and the regularization.  

In the new MAE loss function, $||W_2^T MW_2-I||_{F}^{2}$ defines a functionally independent regularization to force the column vectors of $W_2$  to be mutually orthogonal with respect to the matrix $M$, namely 
\begin{center}
  $ w_i^TMw_j=\left \{
        \begin{array}{lr}
           0, i \ne j & \\ 
           1, i = j &
        \end{array}
  \right.$
\end{center}
where $w_i$ is the $i$-$th$ column vector of $W_2$. The column vectors are mutually orthogonalised and normalised when $M=I$. When  $M=diag\left\{ {{\left( w_{i}^{T}{{w}_{i}} \right)}^{-1}} \right\}$, the column vectors will be mutually orthogonal but not normalised, which is the setting we used in the upcoming discussion.  

According to the structure in Figure \ref{fig1}, the output $\hat{X}$ of the decoder can be reconstructed from the hidden layer features $Q={[{q}_{1},{q}_{2},\cdots,{q}_{m}]}^{T}$ as
\begin{equation}
  \begin{split}
    \hat{X}=s_g({{W}_{2}}Q+{{b}_{2}})=s_g(\left[ \begin{matrix}
      {{w}_{1}} & {{w}_{2}} & \cdots  & {{w}_{m}}  \\
   \end{matrix} \right]\left[ \begin{matrix}
      {{q}_{1}}  \\
      {{q}_{2}}  \\
      \vdots   \\
      {{q}_{m}}  \\
   \end{matrix} \right]+{{b}_{2}})=s_g(\underset{i=1}{\overset{m}{\mathop \sum }}\,{{w}_{i}}{{q}_{i}}+{{b}_{2}})   
  \label{eq11}
  \end{split}
\end{equation}
where $m$ is the dimension of the hidden layer. A sigmoid activation function, $s_g$, was used in the experimental studies (in Section 4) in order to reconstruct the datasets which have a value between 0 and 1. 

It can be observed that $w_i$ has the same size as $\hat{X}$ and it determines how the latent variable $q_i$ was distributed onto each element of $\hat{X}$ when it was reconstructed. Mutually orthogonal $w_i$'s mean that each of  the latent variables provides an independent source of variation. 

The training of MAE can be summarised in the following procedures: Start with defining the loss function $L$ in Eq.\ref{eq6}. Then initialise  the parameters $\theta$  of the neural network. Update the parameters $\theta$ through training by mini-batch gradient descent (MBGD) \cite{ref67} until a stop criterion is satisfied.

\subsection{Properties of MAE}
\label{subsec3.3}
It can be observed that the MAE model learns meaningful features by introducing the functionally independent regularization. Unlike other autoencoders, which constrain weight matrix $W_1$, MAE applies a constraint to the readout matrix $W_2$. The $i$-$th$ column of matrix $W_2$ determines how the $i$-$th$ hidden layer node $q_j$ contributes to the reconstruction of input. Hence the mutual orthogonalisation of $W_2$ disentangles the factors of variation. It will be shown in the experiment section that this modification significantly improves the quality of learned representations.

\subsection{Extensions of MAE} 
\label{subsec3.4}
Combining the functionally independent constraint with other regularization methods produces other new variations of autoencoder. 

\subsubsection{Extension to DAE} 
\label{subsec3.4.1}
The classic denoising autoencoder can be extended as a modal denoising autoencoder (\textbf{MDAE}) by optimising the following loss function.
\begin{equation}
  \begin{split}
    {{L}_{MDAE}}=\left\| x-g\left( f\left( \tilde{x}\left( \alpha  \right) \right) \right) \right\|_{{}}^{2}+{{\lambda }_{m}}\left\| W_{2}^{T}M{{W}_{2}}-I \right\|_{F}^{2}
  \label{eq8}
  \end{split}
\end{equation}
where $\tilde{x}\left(\alpha\right)$ represents the corrupted input with a noisy level $\alpha$.

\subsubsection{Extension to CAE} 
\label{subsec3.4.2}
Similarly, the CAE loss can be extended as a modal  contractive autoencoder (\textbf{MCAE}) loss as
\begin{equation}
  \begin{split}
    {{L}_{MCAE}}=\left\| x-g\left( f\left( x \right) \right) \right\|_{{}}^{2}+{{\lambda }_{c}}\left\| J\left( x \right) \right\|_{F}^{2}+{{\lambda }_{m}}\left\| W_{2}^{T}M{{W}_{2}}-I \right\|_{F}^{2} 	
  \label{eq9}
  \end{split}
\end{equation}
where $\lambda_c$ and $\lambda_m$ represent the contractive and modal regularization parameters, respectively.

\subsubsection{Extension to SAE}
\label{subsec3.4.3}
A new modal sparse autoencoder (\textbf{MSAE}) training priciple can be defined by the following loss function
\begin{equation}
  \begin{split}
    {{L}_{MSAE}}=\left\| x-g\left( f\left( x \right) \right) \right\|_{{}}^{2}+{{\lambda }_{s}}\left( \left\| {{W}_{1}} \right\|_{F}^{2}\text{+}\left\| {{W}_{2}} \right\|_{F}^{2} \right)+{{\lambda }_{m}}\left\| W_{2}^{T}M{{W}_{2}}-I \right\|_{F}^{2}	
  \label{eq10}
  \end{split}
\end{equation}
where $\lambda_s$ and $\lambda_m$ represent the sparse and modal regularization parameters, respectively. 

The sparse autoencoder may also be realised using a Kullback-Liebler divergence regularization term as ${{\lambda }_{s}}KL\left( \left.{\hat{\rho}}\right|\rho\right)$, where $\hat{\rho}$ represents the activation rate of the hidden neurons, $\rho$ represents a small probability of a the binomial distribution\cite{ref52}.

\section{Experiments}
\label{sec4}
The performance of the proposed MAE model was intensively evaluated on the MNIST dataset \cite{ref53}  and compared on more challenging tasks MNIST variations \cite{ref54} and USPS \cite{ref65}. To highlight the effect of the new regularization, the simplest model structure (MAE+NN), which consists of a single hidden layer MAE followed by a single layer neural network classifier with a softmax activation function, was used. The classification error rate was used as the metric to assess the new model. It is worthy to emphasise that the performance, which is shown in this section, can be significantly improved by stacking multiple MAEs. However, the improved performance can be affected by the settings of multiple hyper-parameters and the more complicated training processes. The multi-sourced performance improvement may not fully come from the newly introduced regularization and this makes the analysis complicated. Therefore, the simplest model structure was used when we undertook the experimental assessment.   

The experiment firstly compares the performance of the proposed MAE models with different values of $\lambda_m$ against BAE to show the efficiency of the new regularisation. The MAE is then compared with the mainstream denoising autoencoder (DAE), contractive autoencoder (CAE), and sparse autoencoder (SAE) on more challenging cases where the MNIST were edited by rotation (rot), corrupted by random noise (bg-rand), corrupted by background image (bg-img) and the combination (bg-img-rot). Finally, we will show that the combination of MAE with the mainstream AEs can improve the performance further.

The MNIST dataset is a well-known database of handwritten digits, which contains a training set of 60,000 examples and a test set of 10,000 samples. Each problem is divided into a training, validation and test set (10000, 10000, 50000 examples respectively). A simple single hidden layer MAE with 8 hidden layer neurons was used as the model. We tried to keep the model as simple as possible to adequately show the effect of the new functionally independent regularization. Please refer to the supplementary materials for the detailed experimental settings.

\begin{figure}[H]
  \centering
  \includegraphics[width = 0.5\linewidth]{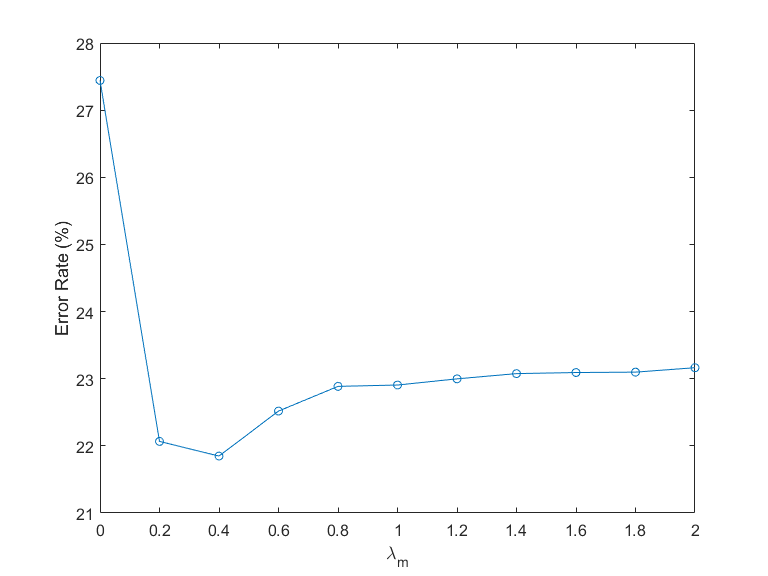}
  \caption{Effect of the regularization of MAE. The comparative results of classification loss on the MNIST dataset between MAE and BAE. It describes how classification loss changes as the scale parameter $\lambda_m$ changes.
  }  
  \label{fig3}
\end{figure}

Classification error rates (in percent) of the MAE+NN model with different values of $\lambda_m$ are shown in Figure \ref{fig3}, which clearly shows how the classification error rates of the MAE change with the effect of the new regularization. It can be observed that the MAE model produced a much smaller classification error than BAE (when $\lambda = 0$) did. Additionally, the performance of the MAE was stable in a large range of $\lambda_m$, though MAE had better performance when the reconstruction loss and the regularization term were well balanced. 

\begin{figure}[H]
  \centering  
  \subfigure[]{ 
  \includegraphics[width = 0.45\linewidth]{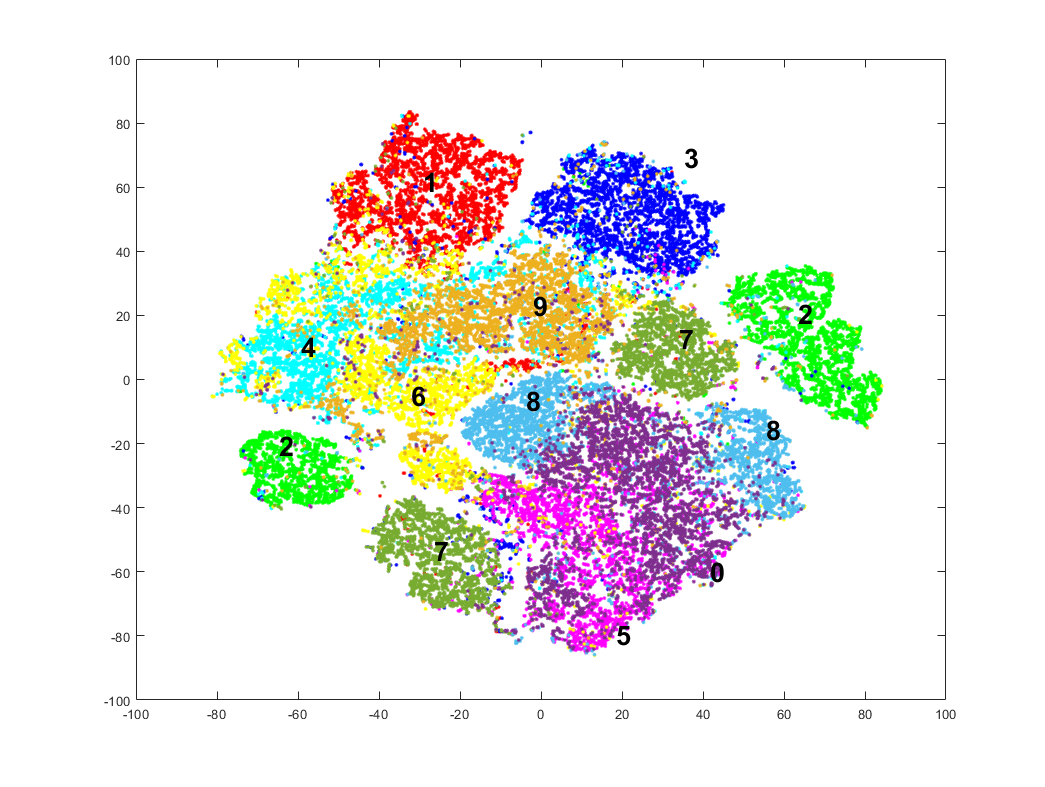}
  }
  \quad
  \subfigure[]{
  \includegraphics[width = 0.45\linewidth]{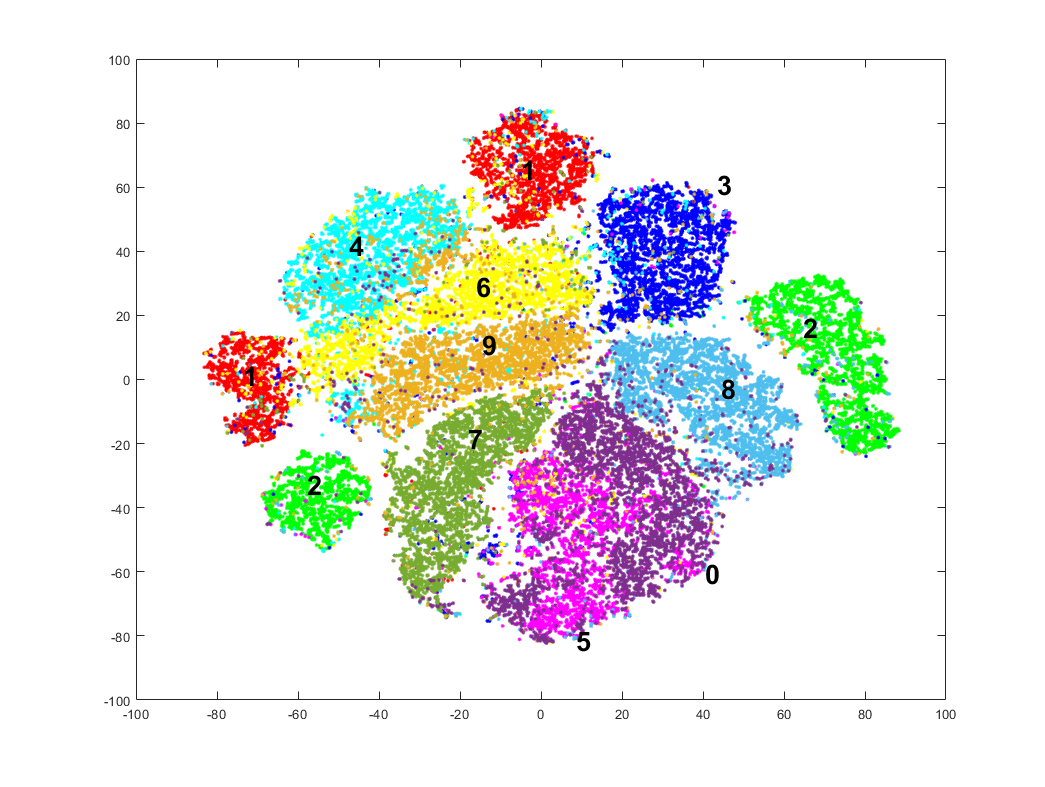}
  }
  \caption{Visualization of clustering results on the MNIST dataset. Different colors mark different clusters according to ground-truth label. (a) The clustering performance of BAE. (b) The clustering performance of MAE.}  
  \label{fig4}
\end{figure}

\begin{figure}
  \centering  
  \includegraphics[width = 0.5\linewidth]{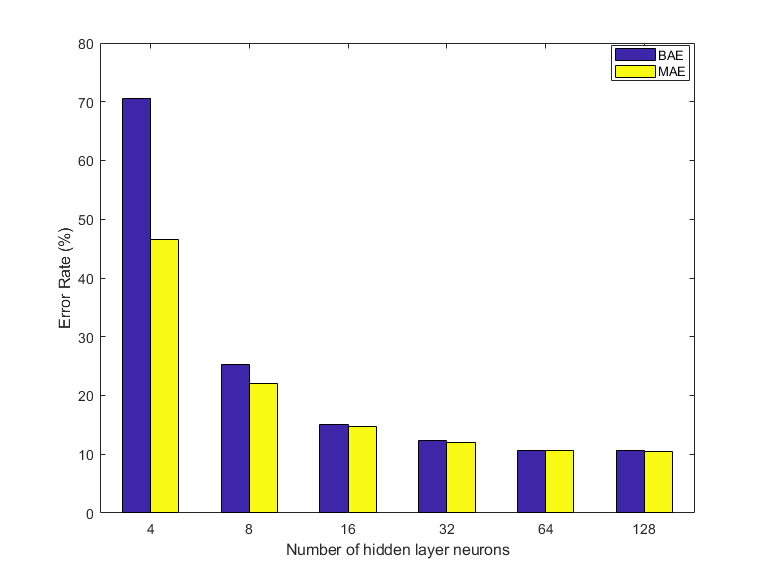}
  \caption{Performance of MAE under different numbers of hidden neurons.The comparative results of classification loss on the MNIST dataset between MAE and BAE. It describes how classification loss changes as the size of the hidden layer varies.
  }  
  \label{fig5}
\end{figure}

To better verify our arguments, we also conducted one more qualitative experiment on the MNIST dataset. The latent variables obtained via BAE and MAE are visualised in Figure \ref{fig4} via t-SNE \cite{ref55} . As shown, MAE provides a  more compact embedding compared to the standard BAE. The numbers are better separated based on the MAE extracted features.
Next, the performances of MAEs with different hidden layer sizes were also studied. Figure \ref{fig5}  shows that under all settings, MAE has a better performance than BAE. However, the absolute difference of the model performances decreases with the increase of the hidden layer neurons. It is easy to understand that MAE extracted more useful information than BAE when the hidden layer is small and produced a much smaller classification error. However, the improvement became less significant when the hidden layer increased. This is because both MAE and BAE can extract much richer information from the original data. 

The new MAE was finally compared with mainstream DAE, CAE, and SAE under the same settings (autoencoders with the same hidden layer size followed by the same classifier, all experiments start from the same initialization of the NNs). All the models were fine-tuned to the best performance. The code for all these experiments was adpated from the MATLAB Toolbox \cite{ref57}. The performance of each model is shown in Table \ref{tab1} where the best performer is shown in bold. The first column gives the baseline of the classification error rate, that is, the error rate given by a single layer neural network (NN) with the raw data as input. It can be observed that MAE appears to achieve a performance superior or comparable to the best other models on most problems. Integrating the new modal regularization into DAE, CAE, and SAE can further improve the performance.

\begin{table}
  \caption{Comparison of MAE with other AEs on the benchmarks$^{*}$ }
  \label{tab1}
  \centering
  \scalebox{0.8}{
  \begin{tabular}{cccccccccc}
    \toprule
    \multirowcell{2}{Dataset} & \multirowcell{2}{NN} & \multirowcell{2}{BAE} & \multirowcell{2}{DAE \\ $(\alpha)$} &  \multirowcell{2}{ CAE \\ $(\lambda_c)$} & \multirowcell{2}{ SAE \\ $(\lambda_s)$} & \multirowcell{2}{\textbf{MAE} \\ $(\lambda_m)$} & \multirowcell{2}{ \textbf{MDAE} \\$(\alpha,\lambda_m)$} & \multirowcell{2}{\textbf{MCAE} \\$(\lambda_c,\lambda_m)$} & \multirowcell{2}{\textbf{MSAE} \\$(\lambda_s,\lambda_m)$} \\
          &      &     &   &   &  &   &  &  & \\
    \midrule
    \multirowcell{2}{MNIST}    &  \multirowcell{2}{52.47}    &  \multirowcell{2}{25.27}  &  \multirowcell{2}{24.88 \\ $(5\%)$} &  \multirowcell{2}{24.77 \\(0.001)}  & \multirowcell{2}{23.71 \\(0.1)}   &   \multirowcell{2}{22.11 \\(0.4) }  &  \multirowcell{2}{\textbf{21.81} \\ $(5\%,0.4)$} &  \multirowcell{2}{ 22.03 \\(0.001,0.4)}  &  \multirowcell{2}{22.93 \\$(0.1,0.4)$}    \\
    &      &     &   &   &  &   &  &  & \\
    \cmidrule{4-10}
    \multirowcell{2}{bg-rand}    &  \multirowcell{2}{82.25}    &  \multirowcell{2}{29.74}   & 29.80 & 29.86 &  28.16   &  29.31   & 29.11     & 28.35 & \textbf{27.60}   \\
    &      &     &  $(1\%)$  & (0.001)  & (0.05) & (0.001)  & $(1\%,0.001)$ & (0.001,0.001) & (0.05,0.001) \\
    \cmidrule{4-10}
    \multirowcell{2}{bg-img}   &  \multirowcell{2}{78.54}    &  \multirowcell{2}{62.08}  & 62.34 & 62.26 &  62.75 &  \textbf{61.43}  & 61.88  & 61.45  &  62.02    \\
    &      &     &  $(1\%)$  & (0.001)  & (0.001) & (0.1)  & $(1\%,0.1)$ & (0.001,0.1) & (0.001,0.1) \\
    \cmidrule{4-10}
    \multirowcell{2}{rot}   &  \multirowcell{2}{77.41}    &  \multirowcell{2}{89.01}  & 88.32 & 89.04 &  55.82   &  54.45  & 54.33     & 54.28 &  \textbf{54.01}    \\
    &      &     &  $(1\%)$  & (0.001)  & (0.1) & (0.1)  & $(1\%,0.1)$ & (0.001,0.1) & (0.1,0.1) \\
    \cmidrule{4-10}
    \multirowcell{2}{bg-img-rot}   &  \multirowcell{2}{84.58}   &  \multirowcell{2}{77.32}  & 79.87 &77.81  &  83.71   & 77.22 & 79.43 &  \textbf{76.83}  &  81.78    \\
    &      &     &  $(10\%)$  & (0.01)  & (0.05) & (0.0005)  & $(10\%,0.0005)$ & (0.01,0.0005) & (0.05,0.0005) \\
    \cmidrule{4-10}
    \multirowcell{2}{USPS}   &  \multirowcell{2}{71.92}    &  \multirowcell{2}{89.25}  & 89.90 & 89.26  &  46.52   &  52.23  & 52.41 & 51.28   &  \textbf{43.79}    \\
    &      &     &  $(1\%)$  & (0.001)  & (0.1) & (0.7)  & $(1\%,0.7)$ & (0.001,0.7) & (0.1,0.7) \\
    \bottomrule
    \multicolumn{8}{l}{$^{*}$Hyper-parameters were the same as the definition in subsection 3.4} & & \\
  \end{tabular}}
\end{table}

In sum, the new MAE provided good performance and worked well under different settings. The obtained functionally independent features reflect the basic patterns underlying the input data. This indicates that the constraint to the readout weights can significantly improve the learned representations, which performed surprisingly well in the subsequent classification tasks. 

\section{Conclusions}
\label{sec5}
We have introduced a very simple training principle for autoencoders, based on the objective of extracting functionally independent features from the input. This was inspired by the modal analysis in mechanical and civil engineering where the modal shapes are mutually orthogonal. A series of image classification experiments have been conducted to evaluate the MAE model. The empirical results support the following conclusions: introducing a mutually orthogonal (functionally independent) regularization to the decoder weights helps to disentangle the factor of variations and to capture interesting structures in the input distribution. This in turn leads to intermediate representations much better suited for subsequent learning tasks. Comparative studies show that MAE performs better than mainstream autoencoders in most cases. It is easy to integrate the new functionally independent regularization with other autoencoders to give the learned representations comprehensive properties. This principle could be a promising way to train and stack autoencoders to initialise a deep neural network. 

\section*{Broader Impact}
DNNs try to improve AI by exploring the meaningful features automatically. Inspired by the engineering method - modal analysis, the modal autoencoder improves the representation learning and extracts more meaningful features. The impacts of the results are twofold: The results of this research will significantly advance knowledge and understanding on the hidden layer nodes of deep neural networks and makes neural network models more interpretable. The new training principle can significantly improve the performance of neural networks in object recognition, natural language processing, signal processing and so on. Integrating machine learning theory and engineering and neural science, MAE may push the direct application of AE in innovative research and help translate the research findings into engineering practice. The MAE was inspired by linear modal analysis in mechanical engineering. However, MAE provides a powerful tool for extracting nonlinear features from experimental recordings. This may significantly simplify experimental modal analysis in mechanical engineering and advance the knowledge of nonlinear modal analysis. In our current research, MAE is also being used to study brain activities by extracting the fundamental brain dynamic modes from electroencephalogram (EEG) signal. Application of MAE, instead of cross-correlation methods, will generate crucial insights into brain functional connectivity which will make it possible to provide a complete picture of whole-brain activities rather than paired relationships. The related research may have important impacts on brain-computer interface and neurorehabilitation.

\begin{ack}
Authors gratefully acknowledge the support from the National Natural Science Foundation of China (Grant No. 61876015), the Beijing Natural Science Foundation, China (Grant No. 4202040), and Science and Technology Innovation 2030 Major Program of China (Grant No. 2018AAA001400).  
\end{ack}

\medskip
\bibliographystyle{ieeetr}
\bibliography{myreference}

\begin{thebibliography}{10}

\bibitem{ref50}
Y.~{Bengio}, A.~{Courville}, and P.~{Vincent}, ``Representation learning: A
  review and new perspectives,'' {\em IEEE Transactions on Pattern Analysis and
  Machine Intelligence}, vol.~35, pp.~1798--1828, Aug 2013.

\bibitem{ref66}
T.~Xiao, Y.-H. Tsai, K.~Sohn, M.~Chandraker, and M.-H. Yang, ``Adversarial
  learning of privacy-preserving and task-oriented representations,'' 2019.

\bibitem{ref59}
S.~Balakrishnama and A.~Ganapathiraju, ``Linear discriminant analysis-a brief
  tutorial,'' vol.~11, 01 1998.

\bibitem{ref60}
U.~Demšar, P.~Harris, C.~Brunsdon, A.~Fotheringham, and S.~Mcloone,
  ``Principal component analysis on spatial data: An overview,'' {\em Annals of
  the Association of American Geographers}, vol.~103, no.~1, pp.~106--128,
  2013.

\bibitem{ref61}
D.~Charte, F.~Charte, S.~Garc{\'{\i}}a, M.~J. del Jesus, and F.~Herrera, ``A
  practical tutorial on autoencoders for nonlinear feature fusion: Taxonomy,
  models, software and guidelines,'' {\em CoRR}, vol.~abs/1801.01586, 2018.

\bibitem{ref51}
P.~Vincent, H.~Larochelle, Y.~Bengio, and P.-A. Manzagol, ``Extracting and
  composing robust features with denoising autoencoders,'' in {\em Proceedings
  of the 25th International Conference on Machine Learning}, (New York, NY,
  USA), pp.~1096--1103, Association for Computing Machinery, 2008.

\bibitem{ref62}
S.~Rifai, P.~Vincent, X.~Muller, X.~Glorot, and Y.~Bengio, ``Contractive
  auto-encoders: Explicit invariance during feature extraction,'' in {\em
  ICML}, 2011.

\bibitem{ref63}
J.~Deng, Z.~Zhang, E.~Marchi, and B.~Schuller, ``Sparse autoencoder-based
  feature transfer learning for speech emotion recognition,'' pp.~511--516, 09
  2013.

\bibitem{ref64}
D.~RUMELHART, G.~HINTON, and R.~WILLIAMS, ``Learning internal representations
  by error propagation,'' in {\em Readings in Cognitive Science} (A.~Collins
  and E.~E. Smith, eds.), pp.~399 -- 421, Morgan Kaufmann, 1988.

\bibitem{ref56}
A.~Pete, ``Modal space - in our own little world,'' {\em SEM Experimental
  Techniques}, Feb 1998.

\bibitem{ref67}
G.~Hinton, N.~Srivastava, and K.~Swersky, ``Neural networks for machine
  learning lecture 6a overview of mini-batch gradient descent,'' {\em Cited
  on}, vol.~14, no.~8, 2012.

\bibitem{ref52}
Q.~{Meng}, D.~{Catchpoole}, D.~{Skillicom}, and P.~J. {Kennedy}, ``Relational
  autoencoder for feature extraction,'' in {\em 2017 International Joint
  Conference on Neural Networks (IJCNN)}, pp.~364--371, May 2017.

\bibitem{ref53}
Y.~{Lecun}, L.~{Bottou}, Y.~{Bengio}, and P.~{Haffner}, ``Gradient-based
  learning applied to document recognition,'' {\em Proceedings of the IEEE},
  vol.~86, pp.~2278--2324, Nov 1998.

\bibitem{ref54}
H.~Larochelle, D.~Erhan, A.~Courville, J.~Bergstra, and Y.~Bengio, ``An
  empirical evaluation of deep architectures on problems with many factors of
  variation,'' in {\em Proceedings of the 24th International Conference on
  Machine Learning}, ICML '07, (New York, NY, USA), pp.~473--480, Association
  for Computing Machinery, 2007.

\bibitem{ref65}
M.~Min, D.~Stanley, Z.~Yuan, A.~Bonner, and Z.~Zhang, ``A deep non-linear
  feature mapping for large-margin knn classification,'' pp.~357--366, 12 2009.

\bibitem{ref55}
L.~v.~d. Maaten and G.~Hinton, ``Visualizing data using t-sne,'' {\em Journal
  of machine learning research}, vol.~9, no.~Nov, pp.~2579--2605, 2008.

\bibitem{ref57}
R.~B. Palm, ``Prediction as a candidate for learning deep hierarchical models
  of data,'' Master's thesis, 2012.

\end{thebibliography}

\end{document}